\documentclass[conference]{IEEEtran}
\IEEEoverridecommandlockouts
\usepackage[compatibility=false, footnotesize]{caption}
\usepackage[
style=ACM-Reference-Format,
sorting=none, 
doi=false,
isbn=false,
url=false,
eprint=true,
style=numeric-comp,
maxcitenames=2,
maxbibnames=2,
backend=biber,
backref=false
]{biblatex}
\usepackage{algorithm}
\usepackage{algpseudocode}
\usepackage{booktabs}

\usepackage{graphicx}
\usepackage{subfig}

\usepackage{makecell}
\usepackage{multirow}
\usepackage{enumitem}
\setlist{nolistsep}
\usepackage{amsmath,amssymb,amsfonts}
\usepackage{graphicx}
\usepackage{textcomp}
\usepackage{xcolor}
\def\BibTeX{{\rm B\kern-.05em{\sc i\kern-.025em b}\kern-.08em
    T\kern-.1667em\lower.7ex\hbox{E}\kern-.125emX}}

\newcommand{\eqn}[1]{Eqn.$\,$(\ref{eqn:#1})}

\def\bea{\begin{eqnarray}}
\def\eea{\end{eqnarray}}
\newcommand{\fig}[1]{Fig. \ref{fig:#1}}

\DeclareMathSizes{10}{9}{7}{5} % Example: 10pt document, math sizes 9, 7, and 5
\allowdisplaybreaks

\begin{document}
\title{A Parallel Alternative for Energy-Efficient Neural Network Training and Inferencing}
%\title{Compression-Induced Communication-Efficient Large Model Training and Inferencing}
\author{
%\IEEEauthorblockN{Anonymous Authors}
%\iffalse 
\IEEEauthorblockN{Sudip K. Seal$^a$, Maksudul Alam$^a$, Jorge Ramirez$^a$, Sajal Dash$^b$ and Hao Lu$^b$}
  $^a$Computer Science and Mathematics Division, $^b$The National Center for Computational Sciences  \\  
  Oak Ridge National Laboratory, USA
%\thanks{This manuscript has been authored by UT-Battelle, LLC under Contract No. DE-AC05-00OR22725 with the U.S. Department of Energy. The United States Government retains and the publisher, by accepting the article for publication, acknowledges that the United States Government retains a non-exclusive, paid-up, irrevocable, worldwide license to publish or reproduce the published form of this manuscript, or allow others to do so, for United States Government purposes. The Department of Energy will provide public access to these results of federally sponsored research in accordance with the DOE Public Access Plan (http://energy.gov/downloads/doe-public-access-plan). This research was sponsored by Oak Ridge National Laboratory’s Laboratory Directed Research and Development program. This research used resources at the Oak Ridge Leadership Computing Facility which is a DOE  Office of Science User Facility.}
%\fi  
}

\maketitle
\begin{abstract}
Energy efficiency of training and inferencing with large neural network models is a critical challenge facing the future of sustainable large-scale machine learning workloads. This paper introduces an alternative strategy, called phantom parallelism, to minimize the net energy consumption of traditional tensor (model) parallelism, the most energy-inefficient component of large neural network training. The approach is presented in the context of feed-forward network architectures as a preliminary, but comprehensive, proof-of-principle study of the proposed methodology. We derive new forward and backward propagation operators for phantom parallelism, implement them as custom autograd operations within an end-to-end phantom parallel training pipeline and compare its parallel performance and energy-efficiency against those of conventional tensor parallel training pipelines. Formal analyses that predict lower bandwidth and FLOP counts are presented with supporting empirical results on up to 256 GPUs that corroborate these gains. Experiments are shown to deliver $\sim 50\%$ reduction in the energy consumed to train FFNs using the proposed phantom parallel approach when compared with conventional tensor parallel methods. Additionally, the proposed approach is shown to train smaller phantom models to the same model loss on smaller GPU counts as larger tensor parallel models on larger GPU counts offering the possibility for even greater energy savings.
%When compared with conventional tensor parallel approaches, empirical experiments are shown to deliver over two orders of magnitude reduction in energy consumption while registering an order of magnitude reduction in the time to train feed forward network models to fixed losses using the proposed approach.
\end{abstract}


\section{Introduction}
\label{sec:intro}

%\subsection{Motivation}      
%\label{sec:motivation}
Large language models (LLMs) are popular examples of super-sized neural network (NN) models used for natural language processing tasks, but a wide array of foundation models for other data modalities have increasingly become popular. 
%Examples include models for texts-to-images~\cite{dalle}, text-to-3D images~\cite{dreamfusion}, images-to-text~\cite{flamingo}, text-to-audio~\cite{audiolm}, text-to-text~\cite{chatgpt}, as well as multi-modal ones~\cite{aghajanyan2022cm3}. 
Training these models, which require weeks or even months, typically on high-performance supercomputers equipped with accelerators (such as graphical processing units or GPUs), costs millions of dollars~\cite{nagrecha2023systems} and consume enormous amounts of energy. 

For example, training GPT-3 (175B parameters) reportedly consumed approximately 1,287 MWh of electricity, equivalent to the annual consumption of 120 US households, and resulted in more than 552 metric tons of CO\textsubscript{2} emissions~\cite{strubell2019energy, patterson2021carbon}. The 176B-parameter BLOOM model training campaign emitted between 24.7 and 50.5 tonnes of CO\textsubscript{2}eq, depending on whether manufacturing emissions were included~\cite{luccioni2023bloom}. Although training is a one-time cost, inference happens continuously at scale; for example, OpenAI’s ChatGPT reportedly served more than 1 billion queries per month as of 2023, leading to enormous aggregate energy consumption. Patterson et al.~\cite{patterson2021carbon} estimate that inference costs eventually outstrip training energy costs by a factor of 2-10$\times$ over the lifetime of a model. As model sizes continue to grow to achieve gains in performance, the energy and carbon footprint associated with both phases become formidable, potentially unsustainable, impacting not only operational costs but also the broader environmental footprint of AI technologies. The sheer sizes of these models, the associated training costs, and the resulting environmental impact necessitate a renewed look into the energy efficiencies of training them. This paper is motivated by the urgency of developing alternative strategies to minimize the monetary and carbon footprint of training and inferencing using large NNs. 

\subsection{Energy Efficiency and Training Parallelism}
\label{sec:enef}
In the context of large-scale training of ML models, data-parallel, pipeline-parallel, and model-parallel methods are the most well-known approaches~\cite{nvidia2020, icml2020}. Data-parallel methods train a model on separate subsets of the training data without any modification to the model architecture. This method is appropriate when the size of the ML model fits in the memory of a single computing host, usually a GPU, and has been shown to scale well with the size of the training data due to its embarrassingly parallel execution style~\cite{dataparallelpaper}. For models that do not fit in the memory of a single accelerator, there are two main alternatives - {\em pipeline} parallelism~\cite{gpipe,pipedream} and {\em tensor} parallelism~\cite{sc21tp}. In pipeline parallelism, groups of whole layers of a large model are partitioned across multiple GPUs (vertical partitioning). In tensor parallelism, each layer is partitioned across multiple GPUs (horizontal partitioning)~\cite{shoeybi2019megatron}. Both styles of parallelism incur multiple rounds of expensive collective communications that are used to periodically aggregate the partitioned information and correctly train the model. 

In practice, most large-scale ML training workloads use all three parallel approaches concurrently to maximize scalability in terms of the training data size (data parallelism) and memory usage (model parallelism), with model parallelism contributing the most to the overall communication costs of any large model training campaign. A significant portion of the net energy consumption of these large-model training workloads can be traced to the computation and communication characteristics of the underlying training algorithms. 
%The energy and communication tradeoffs associated with different parallel training strategies are summarized in terms of their key characteristics in Table~\ref{tab:parallelism-energy}. 
Data parallelism generally offers the best energy efficiency due to minimal communication overhead, while model (tensor) parallelism incurs significant energy costs driven by extensive inter-device synchronization. Pipeline parallelism sits between these two extremes, and its efficiency is highly dependent on batch scheduling and pipeline depth. In this paper, we focus on model parallelism and introduce a new approach that seeks to minimize the energy usage incurred in model parallel training and inferencing.

\subsection{Related Work}
\label{sec:related}
The energy footprint of large neural network training has been widely studied in recent years. Early work by Strubell et al.~\cite{strubell2019energy} highlighted the carbon emissions associated with deep learning models, emphasizing the urgent need for energy-efficient practices. Patterson et al.~\cite{patterson2021carbon} provided a detailed analysis of carbon costs in many major ML models, illustrating that communication overheads can account for a substantial fraction of total energy usage during training.

%Parallel training strategies have been extensively explored to scale model sizes. Data parallelism, where model replicas are trained independently across batches of data, is energy-efficient when the model fits within a single device memory~\cite{goyal2017accurate}. However, for models exceeding device memory, pipeline parallelism~\cite{huang2019gpipe} and tensor (model) parallelism~\cite{shoeybi2019megatron} have been proposed. While effective for scalability, these approaches introduce significant communication overhead, which negatively impacts energy efficiency, as observed in large-scale training efforts such as GPT-3~\cite{brown2020language}.

Efforts to address energy challenges include a variety of approaches. Low-/mixed-precision arithmetic attempts to reduce energy and memory bandwidth while maintaining accuracy through loss scaling~\cite{wang2018training, micikevicius2018mixed}. Model compression techniques attempt to compress the gradients to reduce the inter-process communication overheads~\cite{han2015deep}. Sparsity-aware training reduces net computing overheads by activating or updating a subset of parameters per sample~\cite{evci2020rigging, 10.5555/3586589.3586709}. Sparsity-aware training modifies the model or training loop to introduce or exploit zeros in weights, activations, or gradients, often requiring masking, thresholding, or custom sparse operations and hardware support unlike the proposed method that only modifies the parallelism strategy and could potentially be used in conjunction with sparsity-aware methods for even greater efficiency. Hardware innovations such as Tensor Processing Units (TPUs)~\cite{jouppi2017datacenter} and neuromorphic systems~\cite{davies2018loihi} aim to mitigate energy consumption, but require specialized infrastructure not yet widely accessible.

Unlike hardware-centric approaches, this paper focuses on algorithmic strategies for designing energy-aware parallel training algorithms that reduce computation, communication, and idle hardware time, and complement ongoing power-aware hardware developments while offering immediate solutions for extreme-scale training on existing supercomputing resources. 
%The approach proposed here can also be used in conjunction with existing software strategies highlighted above. 

\subsection{Contributions, Significance and Limitations}
\label{sec:contrib}
This paper makes the following contributions:
\begin{itemize}
    \item A new energy-efficient paradigm, called {\bf phantom parallelism}, for model parallel training is introduced.
    \item A comprehensive formal analysis of its potential for energy savings compared to conventional tensor parallelism is presented. 
    \item New forward and backward operators are derived and implemented as custom autograd operations for execution on distributed parallel architectures.
    \item Performance comparisons of phantom and conventional tensor parallelism in terms of parallel execution characteristics and energy consumption are presented on up to 256 GPUs to confirm the potential benefits.
\end{itemize}

The results presented here are limited to simple FFN architectures. The choice of FFNs is deliberate. It represents a NN architecture that is simple enough for derivations and implementations of non-trivial forward and backward phantom operators needed to test the parallel performance and energy efficiency of the proposed training pipeline and for comparisons with those of tensor parallel training pipelines. Notably, more complex NN models often include FFNs as components within their overall design, and the results reported here for standalone FFNs are equally applicable to those in more complex NN architectures (e.g., a transformer model has a FFN layer after each attention layer).

The authors recognize that further research is required to generalize the methods and results presented here for applicability to more complex NN architectures and caution the reader against assuming that the gains reported here can be translated unmodified in the most general cases. However, in the face of a looming energy crisis from the global adoption of super-sized AI tools and applications, the authors present this work as an initial but significant step in the direction of more energy-conscious at-scale AI/ML training and inferencing.

%The proposed phantom parallel approach offers a scalable and energy-aware parallelism primitive, laying the groundwork for broader integration into transformer models, inference pipelines, and hybrid parallel strategies. \textcolor{red}{Its practical impact is in its potential for orders-of-magnitude  reductions in the energy consumption associated with training large AI/ML models and their subsequent deployments as inference engines which are almost always characterized by unbounded costs in revenue and energy.}

%We introduce the model used to compare and assess energy efficiency in Section~\ref{sec:prelims}. The principles of phantom parallelism, the challenges to implementing the approach and their solutions are presented in Section~\ref{sec:phantom}. Section~\ref{sec:analysis} provides a comprehensive formal analysis of the phantom approach followed by empirical performance results in Section~\ref{sec:perf}. We summarize in Section~\ref{sec:summary}.   
 
\section{Preliminaries}
\label{sec:prelims}
We begin by introducing the energy model used to assess the energy consumption of training a model using regular tensor parallelism (TP) and the proposed phantom parallelism (PP). In this paper, $p$ (without subscript) will refer to the total number of parallel ranks and $p_i$ will refer to the process with rank $i$. Unless otherwise specified, $p$ ranks will implicitly refer to $p$ GPUs (each rank is assigned to a GPU). Additionally, $n$ will be used to denote a measure of the size of a NN model to be trained (e.g., the width of a layer in a FFN). 

\subsection{Energy Model}
\label{sec:emodel}
Let $e(n,p,L)$ denote the energy consumed (say, in Joules) per iteration per layer of a deep NN of size $n$ and depth $L$ trained using $p$ GPUs. An iteration is defined as one forward pass of the pipeline followed by a backward propagation.
%(\textcolor{red}{Q: each data item or the entire data set?}). 
We model this energy, $e(n,p,L)$, as:
\bea\label{eqn:epi}
e(n,p,L)=A\cdot \alpha(n,p,L) + B\cdot \beta(n,p,L)
\eea
where the constants $A$ and $B$ in \eqn{epi} represent the dynamic and static energy consumption rates (in Watts) of the devices, respectively. Since each rank is assigned to a GPU, the constants $A$ and $B$, represent the energy consumption rates (in Watts) when a GPU accelerator is busy (during computation) or idle (during communication), respectively. The parameters $A$ and $B$ are vendor-specific hardware constants where $A>B$. For example, $A\sim 560W$ and $B\sim 90W$ in the \textsc{Frontier} supercomputer. The coefficient $\alpha(n,p,L)$ is the parallel computation cost (in seconds), while the coefficient $\beta(n,p,L)$ is the parallel communication cost (in seconds) incurred per iteration when training the NN model to the final loss, $\lambda$, using $p$ GPUs.  

Let $\nu_\lambda$ denote the number of iterations required to train the model to a final, $\lambda$. Therefore, the net energy consumption, $E_\lambda(n,p,L)$, of training the model to the target loss, $\lambda$, is:
\bea\label{eqn:totale}
E_\lambda(n,p,L)=\nu_\lambda\cdot e(n,p,L)
\eea
For ease of presentation, we will {\em suppress the subscript $\lambda$} with the understanding that all comparisons of energy consumption during model training using TP and PP are made with respect to the same final loss, $\lambda$.

\subsection{FFN and Tensor Parallelism}
\label{sec:ffntp}
\begin{figure}[t]
    \centering
    \includegraphics[scale=0.8]{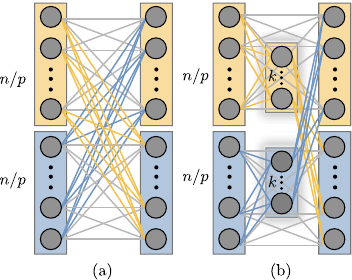}
    \caption{Parallel partitioning of a fully connected single layer FFN ($p=2$).}
    \label{fig:ffntppp}
\end{figure}

Consider a fully connected single layer FFN with input and output layers of width $n$ connected by a weight matrix $\mathbf{W}$ of size $n\times n$ partitioned across $p$ ranks. \fig{ffntppp}(a) illustrates a TP partitioning of such a FFN model across two ranks ($p=2$). The neurons in the local portion of the output layer in a given rank depend on $O(\frac{n^2}{p^2})$ computations involving the local portion of the input layer. In addition, $\frac{n}{p}$ values are received from each of the remaining $(p-1)$ ranks. This incurs an additional computation volume per rank that scales as $O(\frac{n^2}{p^2}(p-1))=O(\frac{n^2}{p})$. 

\begin{figure}[t]
    \centering
    \includegraphics[scale=0.8]{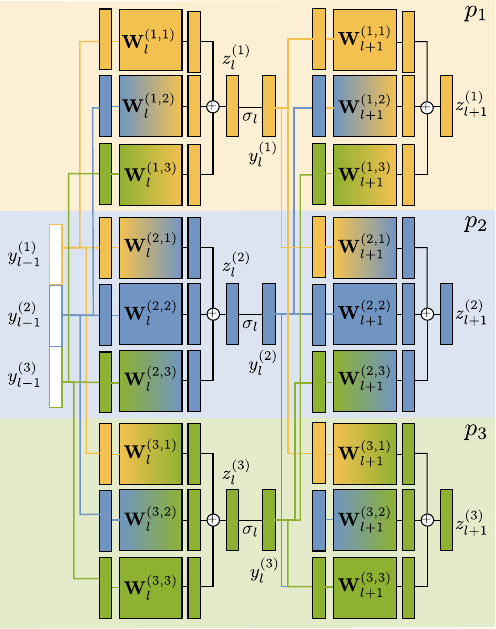}
    \caption{A 3-way TP partition of a FFN with two layers. Mono-colored block computations do not depend on remote information while bi-colored block computations depend on information in corresponding colored ranks. The colored lines indicate inter-process communications.}   
    \label{fig:3waytpffn}
\end{figure}

 The local portion of the global weight matrix $\mathbf{W}$ assigned to a rank $i$ can be thought of as partitioned into $p$ blocks, each of size $[\frac{n}{p}\times \frac{n}{p}]$, which need to be updated locally during training. Of these $p$ blocks, $(p-1)$ depend on remote portions of the global input layer, and one block depends only on the local portion. These $p$ blocks are explicitly shown in \fig{3waytpffn} for a 3-way partitioning ($p=3$) of a FFN with two layers. For each rank $i$, these blocks are labeled $\mathbf{W}_l^{(i,j)}$ where the subscript denotes the layer index and the superscripts $(i,j)$ refer to the local rank $i$ and the remote rank $j$ on which the block computations depend. In \fig{3waytpffn}, mono-colored block computations do not depend on remote information, while bi-colored block computations depend on information in matching colored ranks. It is straightforward to conclude that for a single layer, $(n^2/p^2)$ weights are computed locally and $n^2(p-1)/p^2=O(n^2/p)$ weight computations require communication in {\em each} forward pass of a training iteration. Therefore, the total computation volume to update the global weight matrix $\mathbf{W}_l$ scales as $O(n^2)$ across {\em all} ranks {\em per} layer {\em per} forward pass. The computational cost during a backward pass can also be shown to be of the same complexity \cite{nvidia-megatron-tp, pytorch-colwise-parallel}. Thus, it follows that for a FFN with $L$ layers:
\bea\label{eqn:tpcomp}
\alpha_\tau(n,p,L)&=&L\cdot O(n^2)
\\\label{eqn:tpcomm}
\beta_\tau(n,p,L)&=&L\cdot O(p\log p + n)
\eea
where the subscript $\tau$ is used to denote TP and the collective communications used per TP iteration (forward and backward passes) with their respective costs are listed in the Appendix. Combining \eqn{totale}, \eqn{tpcomp} and \eqn{tpcomm}, it follows that the total energy consumed {\em per} iteration when training a FFN of size $n$ and depth $L$ using $p$ GPUs is:
\bea\label{eqn:tpe}
e_\tau(n,p,L)=A\cdot\alpha_\tau(n,p,L)+ B\cdot\beta_\tau(n,p,L)
\eea
where $\alpha_\tau$ and $\beta_\tau$ are given by \eqn{tpcomp} and \eqn{tpcomm}.
%which will form the baseline for comparison of energy-efficiency with the proposed PP approach described next.
\section{Phantom Parallelism}
\label{sec:phantom}
Consider \eqn{tpe} which represents the total energy consumed per iteration in a $p$-way TP execution. The main idea underlying our proposed approach is to reduce net energy consumption by reducing $\alpha$ and $\beta$ in \eqn{tpe} for the same underlying hardware (same $A$ and $B$, since these are vendor or hardware specific quantities). The proposed approach, called {\bf phantom parallelism (PP)}, attempts to reduce net energy consumption by reducing computation and communication overheads, $\alpha_\tau(n,p,L)$ and $\beta_\tau(n,p,L)$, by introducing additional layers with $k$ neurons where $k\ll  n/p$, as illustrated in \fig{ffntppp}(b) for $p=2$. These additional layers will be referred to as {\em phantom layers} and the $k$ neurons in a phantom layer will be referred to as {\em ghost neurons} to highlight the fact that these layers and neurons are absent in the baseline TP model which the PP model attempts to mimic in terms of performance (training loss) but with less consumption of total energy. It will be shown that the inclusion of phantom layers has the net effect of reducing the total number of computations as well as reducing the net communication overhead.

\subsection{FFN and Phantom Parallelism}
\begin{figure}[t]
%\begin{figure}[thbp]
    \centering
    \includegraphics[scale=0.8]{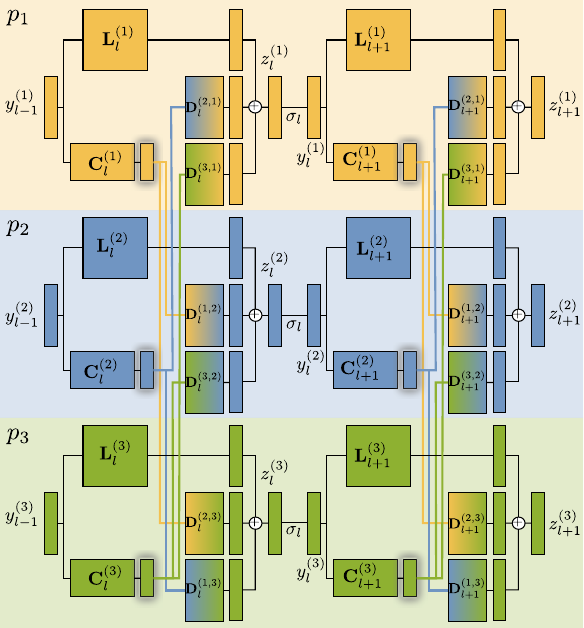}
    \caption{PP components and operations between layer $\ell-1$ and layer $\ell+1$ with $p=3$. Each thick solid-bordered box represents the weight matrix for a specific partition. Dashed boxes denote individual processors, encapsulating their associated weights and intermediate variables. Colored thick lines indicate inter-processor communication.}
    \label{fig:pp-block-3}
\end{figure}

Consider a TP partitioning of a FFN in which the $n$-width input layer is sharded (partitioned) into $p$ partitions, each of width $\frac{n}{p}$. Consider a rank $i$ that is responsible for updating the $\frac{n}{p}$ neurons in the output layer. In a phantom parallel execution, an intermediate phantom layer consisting of $k\ll n/p$ ghost neurons is introduced between the input and output layers in every rank $p_i$ for $0\leq i<p$.
%%
%\begin{figure}[htbp]
\begin{figure}[thbp]
  \centering
  \includegraphics[scale=0.85]{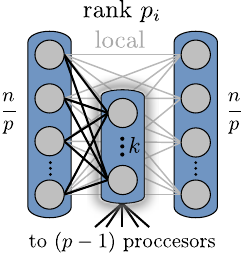}
  \caption{Illustration of a phantom layer of width $k$ inserted between the local $n/p$-width portions of the input and output layers on rank $i$ in a single layer of an FNN of width $n$ partitioned across $p$ processes.}
  \label{fig:ffn_partitioning}
\end{figure}
The intuition behind the PP approach is to compress the information in the local portion of the input layer into a smaller phantom layer before communicating it to the remote ranks (see  \fig{ffn_partitioning}). In a receiving rank, the $(p-1)$ received phantom layers are first locally decompressed to a $\frac{n}{p}$ layer and then used to update the local output layer. These PP operations are explicitly illustrated in \fig{pp-block-3} for a 3-way partitioning ($p=3$) of a FFN with two layers in which the blocks $\mathbf{L}_l^{(i)}$ refers to the weight matrix that connects the local portion of the input layer to the local portion of the output layer, $\mathbf{C}_l^{(i)}$ refers to the computations that compress the local input layer in rank $i$ to a phantom layer with $k$ ghost neurons and  $\mathbf{D}_l^{(j,i)}$ refers to the computations in rank $i$ to decompress the phantom layer received from rank $j$. Compare this figure with its TP counterpart in \fig{3waytpffn}. Although somewhat counterintuitive, we will show next that the total PP computations across all ranks are smaller than the total TP computations for fixed $n$, $L$, and $p$ with the right choices of the number of ghost neurons $k$.

Analogous to TP execution, the energy consumed per iteration in a $p$-way PP execution with $k$ ghost neurons in each phantom layer is modeled as:
\bea\label{eqn:ppe}
e_\pi(n,p,L,k)=A\cdot \alpha_\pi(n,p,L,k) + B\cdot \beta_\pi(n,p,L,k)
\eea
where the subscript $\pi$ refers to PP. As shown in the next section (see \eqn{fppcomp} and \eqn{bppcomp}), the computation cost per layer per process of a $p$-way PP execution with a $k$-width phantom layer scales as $O\left(\tfrac{n^2}{p^2}+kn\right)$. In other words, the total cost of PP computation across all $p$ processes and across all $L$ layers, denoted by $\alpha_\pi(n,p,k,L)$, scales as $L\cdot O\left(\tfrac{n^2}{p}+knp\right)$, in contrast to the total cost of TP computation across all $p$ processes, denoted by $\alpha_\tau(n,p,L)$, which scales as $O(Ln^2)$. It follows, therefore, that:
\bea\label{eqn:compineq}
\alpha_\pi(n,p,k,L) < \alpha_\tau(n,p,L) 
\eea
when 
\bea\label{eqn:kbound}
k< \dfrac{n}{p}\left(1-\dfrac{1}{p}\right)< \dfrac{n}{p}%\text{ for large $p$}
\eea
In the rest of this paper, we assume that the number of ghost neurons or the width $k$ of a phantom layer is always chosen so that it satisfies the above inequality. 

%The communication costs for all the collective communications are discussed in the appendix. 
Using \eqn{kbound}, the message sizes in Table~\ref{tab:commstppp} and Table~\ref{tab:comm-models} and the communication cost modeled in \eqn{ppcollective}, it can be shown that the computation cost per iteration, $\beta_\pi(n,p,k,L)$, in a $p$-way PP execution with $k$ ghost neurons per phantom layer is less than the computation cost per iteration, $\beta_\tau(n,p,L)$, in a TP execution. In other words:
\bea\label{eqn:commineq}
\beta_\pi(n,p,k,L) < \beta_\tau(n,p,L)
\eea
when $k<n/p$. We will prove this inequality in the next section. Since $A$ and $B$ are constants and $A>B$, it follows from \eqn{tpe}, \eqn{ppe}, \eqn{compineq} and \eqn{commineq} that:
\bea\label{eqn:pplttp}
e_\pi(n,p,k,L) < e_\tau(n,p,L)
\eea
implying that the {\bf energy consumed per iteration} in a PP execution is {\bf less} than that consumed in a TP execution for fixed $p$, $n$ and $L$ when $k < n/p$.
%$k<\tfrac{n}{p}\left(1-\tfrac{1}{p}\right)$.

%Note that \eqn{commineq} {\bf does not} imply that $E_\pi(n,p,k,L) < E_\tau(n,p,L)$ since $E_\pi(n,p,k,L)=\nu_\pi\cdot e_\pi(n,p,k,L)$ and $E_\tau(n,p,k,L)=\nu_\tau\cdot e_\tau(n,p,k,L)$ (see \eqn{totale}) and the number of PP and TP iterations ($\nu_\pi$ and $\nu_\tau$, respectively) needed to train a model to a fixed loss, $\lambda$, are not guaranteed to be equal. In Sec.~\ref{sec:perf}, we empirically explore the total energy consumed by the two parallel approaches to reach a {\em fixed} loss.
\section{Phantom Parallel Operations}
\label{sec:ppoperators}
For ease of presentation, we assume that the input, output, and hidden layers of the FFN each have width $n$. A general FFN model may have varying layer widths $\{n_1, n_2,\cdots, n_L\}$ but for the discussion below, $n$ can be viewed as $n=\max\{n_1, n_2,\cdots, n_L\}$. Similarly, we assume $k=\max\{k_1, k_2,\cdots, k_L\}$ without loss of generality.

\subsection{Forward Propagation}
\label{sec:forward}
Consider a $p$-way PP training of the above FNN that models a function $y(x): \mathbb{R}^{n} \to \mathbb{R}^{n}$. The $l$-th layer receives as input the vector $y_{l-1} \in \mathbb{R}^n$ and outputs $y_{l} \in \mathbb{R}^{n}$, $l=1,\dots,L$. The input to the FFN is $y_0 = x$. Each layer is partitioned across $p$ ranks, with rank $p_j$  storing the following components: %described in Section \ref{sec:phantom}: 
\begin{itemize}
    \item a matrix $\mathbf{L}^{(j)}_l$ of size $[\frac{n}{p} \times \frac{n}{p}]$ for the local update,
    \item a compressor $ \mathbf{C}_{l}^{(j)}$ of size $[k \times \frac{n}{p}]$,
    \item $(p-1)$ decompressors $\mathbf{D}_{l}^{(i,j)}$, each of size $[\frac{n}{p} \times k]$ for $i=1,\dots,p$, $i \neq j$.
\end{itemize}
These components are shown in \fig{pp-block-3} for $p=3$. The output $y_l^{(j)}$ in rank $p_j$ is written in terms of its preactivation $z_l^{(j)}$ as:
\begin{equation}\label{eqn:ylq}
 \begin{split}
     y_l^{(j)}  
 &=  \sigma_l \left(b_l^{(j)} + z_l^{(j)}\right),\\ 
 z_l^{(j)} &= \mathbf{L}^{(j)}_l y_{l-1}^{(j)} + \sum_{i\neq j}  \mathbf{D}_{l}^{(i,j)} \underbrace{\mathbf{C}_{l}^{(i)} y_{l-1}^{(i)}}_{g_l^{(i)}}
 \end{split}
\end{equation}
for $l=1,\dots,L$, $j=1,\dots,p$. Here, $y_l^{(j)}$ is of size $\tfrac{n}{p}\times 1$. In Eq. \ref{eqn:ylq}, $\sigma_l$ is the activation function, $b_l^{(j)} \in \mathbb{R}^{n/p}$ are the corresponding bias vectors, and $y_0^{(j)}$ is the component of the input $x$ that is sent to the $j$-th processor.

\subsubsection*{Parallel Complexity}
\label{sec:fcomplexity}
Each rank $p_j$ performs the following operations during each forward pass of an iteration (see \eqn{ylq}):
\begin{itemize}
    \item {\bf Local update:} In this operation, $\mathbf{L}^{(j)}_l y_{l-1}^{(j)}$  is computed. Since $\mathbf{L}^{(j)}_l$ is of size $[\frac{n}{p} \times \frac{n}{p}]$ and $y_{l-1}^{(j)}$ is of size $[\frac{n}{p} \times 1]$, this matrix-vector multiplication costs $O(n^2/p^2)$.
    \item {\bf Compression:} In this operation, $\mathbf{C}_{l}^{(j)} y_{l-1}^{(j)}$ is computed where $\mathbf{C}_{l}^{(j)}$ is of size $[k \times \frac{n}{p}]$ and $y_{l-1}^{(j)}$ of size $[\frac{n}{p} \times 1]$ are locally available in $p_j$. The computation cost of this matrix-vector multiplication costs $O(kn/p)$ and yields a phantom layer, $g_l^{(j)}$, of size $[k\times 1]$ with $k$ ghost neurons.
    \item {\bf Communication:} Each rank communicates this phantom layer to the remaining $(p-1)$ processes, which results in an \texttt{All-Gather} operation with message size $k$. Using \eqn{ppcollective} and Table~\ref{tab:comm-models}, the cost of this communication step is $O(\log p + k)$.
    \item {\bf Decompression:} The $k$-width phantom layers, $g_l^{(i)}$, received from the ranks $p_i$, $i\neq j$, are then locally decompressed using $(p-1)$ matrix-vector computations $\mathbf{D}_{l}^{(i,j)}g_l^{(i)}$. The cost of this operation is $(p-1)O(\frac{n}{p}k)=O(kn)$ since each $\mathbf{D}_{l}^{(i,j)}$ is of size $[\frac{n}{p} \times k]$. 
    \item {\bf Remote update:} Each decompression of a remote ghost layer results in a vector of size $[\frac{n}{p}\times 1]$. These $(p-1)$ $\frac{n}{p}$-width layers are then added to the $\frac{n}{p}$-width layer that resulted from the local update operation. The computation cost of this operation is $O(\frac{n}{p})$.
\end{itemize}
In summary, the net computation and communication costs per layer are $O(n^2/p^2+kn/p)$ and $O(\log p + kb)$, respectively. Therefore, the net computation and communication costs for a single forward pass of a PP pipeline with $k$ ghost neurons per phantom layer to train a $n$-width FFN with $L$ layers across all $p$ ranks are:
\bea
\label{eqn:fppcomp}
 \overrightarrow{\alpha_\pi}(n,p,k,L)&=&L\cdot O(n^2/p+knp)
\\\label{eqn:fppcomm}
\overrightarrow{\beta_\pi}(n,p,k,L)&=&L\cdot O(p\log p + kp)   
\eea
where the arrow indicates forward propagation in an iteration.

%\vspace{-2em}
\subsection{Backward Propagation}
\label{sec:backward}
The output $y_L$ of the full model is obtained by concatenating the outputs $y_L^{(1)},\dots,y_L^{(p)}$ from all $p$ ranks. Each of these outputs is only locally compared with the sharded component $y^{(j)}$ of the data simply denoted by $y$. Let $\mathcal{L}(y_L,y;\theta)$ denote the loss function by comparing the output $y_L(x)$ of the model on some input $x$ with the true output value $y$ as a function of the vector $\theta$ of weights and biases. We will assume that, as in the case of mean-square error or cross-entropy, the loss $\mathcal{L}$ is additive across processors, namely it can be written as
\begin{equation}
    \mathcal{L}(y_L,y;\theta) = \sum_{j =1}^p \mathcal{L}^{(j)}(y_L^{(j)},y^{(j)};\theta).
\end{equation}
where $\mathcal{L}^{(j)}$ denotes the part of the loss function $\mathcal{L}$ that is evaluated from the components of $y_L$ and $y$ stored in the $j$-rank.

Denote the local error with respect to pre-activation $z_l^{(j)}$ for layer $l$ in processor $j$ by $\delta_l^{(j)}$,
\begin{equation}\label{eqn:deltaqqp}
    \delta_l^{(j)} = \frac{\partial \mathcal{L}}{\partial z_l^{(j)}} =  \sum_{i=1}^p \frac{\partial \mathcal{L}^{(i)}}{\partial z_l^{(j)}} = 
    \sum_{i=1}^p \delta_l^{(j,i)}.
\end{equation}
Namely, $\delta_l^{(j,i)} = \partial \mathcal{L}^{(i)}/\partial z_l^{(j)}$ is the contribution in rank $j$ to the $i$-th component of the error in the loss function. For the last layer $\delta_L^{(j,i)} = 0$ for $j \neq i$, and
\begin{equation}\label{eqn:deltaL}
    \delta_L^{(j)} = \delta_L^{(j,j)} =   \frac{\partial \mathcal{L}^{(j)}}{\partial z_L^{(j)}} = \frac{\partial \mathcal{L}^{(j)}}{\partial y_L^{(j)}} \sigma_L'(b_L^{(j)} + z_L^{(j)}) 
\end{equation}
where $\sigma_L'$ denotes the derivative of the activation function.
For all other layers, the backpropagation formula follows from \eqn{ylq}:
\begin{equation*}
\begin{split}
   & \delta_{l}^{(j,i)} = \sum_{i'=1}^p 
    \frac{\partial \mathcal{L}^{(i)}}{\partial z_{l+1}^{(i')}}
    \frac{\partial z_{l+1}^{(i')}}{\partial y_{l}^{(j)}}
    \frac{\partial y_{l}^{(j)}}{\partial z_{l}^{(j)}}=\\
    & \left\{
        \left(\mathbf{L}_{l+1}^{(j)}\right)^\top \delta_{l+1}^{(j,i)}  
       + \sum_{i' \neq j} \left(\mathbf{D}_{l+1}^{(j,i')} \mathbf{C}_{l+1}^{(j)}\right)^{\top} \delta_{l+1}^{(i',i)}
    \right\}  \sigma_l'\left(b_l^{(j)} + z_{l}^{(j)}\right),
\end{split}
\end{equation*}
 Summing over $i$ gives the backpropagation formula for the local error in Eq. \ref{eqn:deltaqqp}:

\begin{equation}\label{eqn:backprop}
\begin{split}
    \delta_{l}^{(j)} &=  \left\{
        \left(\mathbf{L}_{l+1}^{(j)}\right)^\top \delta_{l+1}^{(j)} \quad + \right.\\ 
        &\quad \sum_{i' \neq j}  \left( \mathbf{C}_{l+1}^{(j)}\right)^{\top} \underbrace{\left(\mathbf{D}_{l+1}^{(j,i')} \right)^{\top} \delta_{l+1}^{(i')}}_{h_{l+1}^{(i')}}
    \biggl\}  \sigma_l'\left(b_l^{(j)} + z_{l}^{(j)}\right) 
    \end{split}
\end{equation}
%The asymptotic computing cost of \eqn{backprop} is the same as that of the forward pass in \eqn{ylq}, in other words $O\left(\tfrac{n^2}{p^2}+kn\right)$, as can be observed by comparing the two equations. 
Each backpropagation step requires communicating the vectors $h_{l+1}^{(i')}$ from all $i'$ to all processors. Specific gradients with respect to the weights are given by:
\begin{align}
    \frac{\partial \mathcal{L}}{\partial b_l^{(j)}} &= \delta_l^{(j)}, \label{eqn:bpb}\\
    \frac{\partial \mathcal{L}}{\partial \mathbf{L}_l^{(j)}} &= \delta_l^{(j)} \left( y_{l-1}^{(j)} \right)^{\top}, \label{eqn:bpL}\\
    \frac{\partial \mathcal{L}}{\partial \mathbf{C}_l^{(j)}} &=
    \sum_{i \neq j} \left(\mathbf{D}_l^{(j,i)} \right)^\top \delta_l^{(i)}  \left( y_{l-1}^{(j)} \right)^{\top}, \label{eqn:bpC}\\
    \frac{\partial \mathcal{L}}{\partial \mathbf{D}_l^{(i,j)}} &=
    \mathbf{C}_l^{(i)}  y_{l-1}^{(i)} \left(\delta_l^{(j)}  \right)^{\top}. \label{eqn:bpD}
\end{align}
%The cost of computing the gradients in \eqn{bpL}, \eqn{bpC} and \eqn{bpD} are subsumed by that of computing \eqn{bpb} which is the same as that of \eqn{backprop}. Notably, computing these updates do not incur in any additional communication: all of the terms $ \left(\mathbf{D}_l^{(j,i)} \right)^\top \delta_l^{(i)}$ have already been broadcasted when computing \eqn{backprop}, and $\mathbf{C}_l^{(i)}  y_{l-1}^{(i)}$ was sent from $i$ to $j$ in the forward evaluation. In other words, the only communication call required is in the computation of \eqn{backprop} which is implemented using a \texttt{Reduce-Scatter} with cost shown in Table~\ref{tab:comm-models}. Therefore, the total computation cost {\em per process per layer} of a single iteration in a phantom parallel execution costs $O\left(\tfrac{n^2}{p^2}+kn\right)$ where $k$ is the width of the phantom layer and the communication cost {\em per layer} is the cost of one \texttt{All-Gather} and  one \texttt{Reduce-Scatter}. 

\subsubsection*{Parallel Complexity}
\label{sec:bcomplexity}
Each rank $p_j$ performs the following operations during each backward pass of an iteration (see \eqn{bpb}-\eqn{bpD}):
\begin{itemize}

    \item {\bf Local error at the output:} After the forward pass, the component $\mathcal{L}^{(j)}$ of the loss function is computed. The local output error $\delta_L^{(j)}$ is a $[\frac{n}{p} \times 1]$ vector that is computed from \eqn{deltaL}. In the case of MSE loss, the gradient with respect to the output is $\partial \mathcal{L}^{(j)}/\partial y^{(j)}_L =  (y_L^{(j)} - y^{(j)})$ which has a computation cost $O(n/p)$.

    \item{\bf Error compression:} for each layer $l$, rank $p_j$ compresses its errors to obtain $h_{l}^{(j)} = (\mathbf{D}_{l}^{(j,i)} )^{\top} \delta_{l}^{(j)}$ in \eqn{backprop}. This incurs a computation cost of $O(kn/p)$.

    \item{\bf Communication:} In the forward pass, each rank $i$ communicated its phantom layer $g_l^{(i)}$ to $(p-1)$ ranks. Let $g_l^{(i,j)}$, where $i \neq j$, denote the phantom layer received by rank $j$. During backpropagation, each $g_l^{(i,j)}$ has an associated partial derivative $\frac{\partial\mathcal{L}}{\partial g_l^{(i,j)}}$, which must be sent back and aggregated to the originating rank $i$. Specifically, the gradient with respect to $g_l^{(i)}$ is computed as $\frac{\partial\mathcal{L}}{\partial g_l^{(i)}} = \sum_{i \neq j} \frac{\partial\mathcal{L}}{\partial g_l^{(i,j)}}$. Each gradient is of size $k$. A \texttt{Reduce-Scatter} collective operation is used to execute the aggregation operation in parallel. This gradient aggregation and communication process incurs a communication cost of $O(\log p+k)$.
    
    \item {\bf Local errors:} Having received the compressed errors $\{h_{l+1}^{(i)},i=1,\dots,p\}$, the local error $\delta_{l}^{(j)}$ at layer $l$ is computed with the backpropagation formula in \eqn{backprop}. Taking into account the multiplication by the transposes of the local and compression weight matrices, yields a computation cost of $O(n^2/p^2 + (p-1)k n/p)$ which is $O(n^2/p^2 + kn)$. 

    \item{\bf Individual gradients:} For each layer, the component $y_{l-1}^{(j)}$ in rank $p_j$ is used with its own weight matrices and the local error $\delta_l^{(j)}$ to evaluate the gradient matrices in Eqns. (\ref{eqn:bpb}-\ref{eqn:bpD}). In their totality, these incur a cost of up to $O(n^2/p^2 + (p-1) k n/p)$ which is $O(n^2/p^2 + kn)$.
\end{itemize}
%\textcolor{blue}{At the end of each compressed layer, the phantom data, denoted by $g_l^{(i)}$, is transferred to all other GPUs. The broadcasted phantom value received by GPU $j$ is denoted as $g_l^{(i,j)}$, where $i \neq j$. During backpropagation, each $g_l^{(i,j)}$ has an associated partial derivative $\frac{\partial\mathcal{L}}{\partial g_l^{(i,j)}}$, which must be sent back and aggregated to the originating GPU $i$. Specifically, the gradient with respect to $g_l^{(i)}$ is computed as $\frac{\partial\mathcal{L}}{\partial g_l^{(i)}} = \sum_{i \neq j} \frac{\partial\mathcal{L}}{\partial g_l^{(i,j)}}$. Each of the gradient is of size $kb$ where $b$ is the batch size. To execute the aggregation operation in parallel we use the \texttt{Reduce-Scatter} collective operation. This gradient aggregation and communication process incurs a communication cost of $O(kb \log p)$}

It is straightforward to note that summing up the computation and communication costs for the backward propagation yields:
\bea\label{eqn:bppcomp}
\overleftarrow{\alpha_\pi}(n,p,k,L)&=&\overrightarrow{\alpha_\pi}(n,p,k,L)
\\\label{eqn:bppcomm}
\overleftarrow{\beta_\pi}(n,p,k,L)&=&\overrightarrow{\beta_\pi}(n,p,k,L)
\eea
where the left arrow indicates backward propagation in an iteration. Therefore, the forward and backward costs per PP iteration across all $p$ ranks is:
\bea\label{eqn:ppcomp}
\alpha_\pi(n,p,k,L)&=&L\cdot O(n^2/p+knp)
\\\label{eqn:ppcomm}
\beta_\pi(n,p,k,L)&=&L\cdot O(p\log p + kp)
\eea
Contrast this with \eqn{tpcomp} and \eqn{tpcomm} for TP. \eqn{pplttp} follows from \eqn{tpcomp}, \eqn{tpcomm}, \eqn{ppcomp}, \eqn{ppcomm} and the relation $A>B$ where $A$ and $B$ are constants. We suppress the batch size in our analysis, as it is an overall scale factor for both TP and PP and, therefore, ignored in our discussions bearing in mind that performance comparisons of TP and PP training are carried out with fixed batch sizes.

\section{Parallel Implementation}
\label{sec:implementation}
\begin{algorithm}[t]
\caption{Custom AllGather Autograd Function}
\label{alg:custom_autograd}
\begin{small}
    \begin{algorithmic}[1]
\Function{Forward}{inputs}
    \State output $\gets$ list of empty tensors shaped like \texttt{inputs}
    \For{\textbf{each} $i$ \textbf{in} $0$ \textbf{to} \Call{WorldSize}{} $-1$}
        \State output[$i$] $\gets$ \Call{EmptyLike}{inputs}
    \EndFor
    \State \Call{AllGather}{output, inputs}
    \State \Return \Call{Concat}{output, dim=0}
\EndFunction

\Function{Backward}{grad\_outputs}
    \State input\_list $\gets$ \Call{Chunk}{grad\_outputs, \textsc{WorldSize}}
    \State grad\_inputs $\gets$ \Call{EmptyLike}{input\_list[\textsc{Rank}]}
    \State \Call{ReduceScatter}{grad\_inputs, input\_list}
    \State \Return grad\_inputs
\EndFunction
\end{algorithmic}
\end{small}
\end{algorithm}
The PP framework was implemented using PyTorch 2.7 with Python 3.11 and the RCCL communication library on the \textsc{Frontier} supercomputer using a distributed memory model parallel approach in which the components of each layer (local matrix, compressors, and decompressors) were assigned to individual ranks, each mapped to a single GPU, so that each of the $p$ ranks had exclusive access to a single \textsc{Frontier} GPU.

As highlighted in the previous section, the input and output of each phantom model layer are fully sharded and do not require concatenation, regardless of the number of layers which offers {\em a distinct performance advantage} over TP models, in which the outputs of TP layers must be concatenated every two layers following RowWise and ColWise TP stages~\cite{sc21tp}.

Most of the layer operations -- including the local matrix, compressor, and decompressor -- are matrix-vector multiplications, implemented using the \texttt{Linear} layer from PyTorch’s built-in \texttt{torch.nn} package. 
%When the neural network is constructed, PyTorch automatically creates a computational graph that connects the layers and supports automatic gradient computation. 
The local matrix and compressor operate on the $\frac{n}{p}$ sharded input data local to each GPU. In contrast, the decompressor operates on data not local to the GPU. As a result, this non-resident data must be gathered from all other GPUs during the forward pass. Similarly, during the backward pass, the gradients computed at each decompressor must be communicated back to the GPUs that provided the original input during the forward stage. Figure~\ref{fig:pp-block-3} illustrates these operations for a PP execution with three processes (GPUs). When constructing a neural network, PyTorch automatically generates a computational graph that links the layers and inputs to support automatic differentiation. However, if the input to a layer is not resident on the local GPU, the built-in layers in \texttt{torch.nn} cannot be used directly, which poses an implementation challenge. Native \texttt{torch.nn} does not support this parallel operation, and extra care was necessary to support customized \texttt{torch.nn}-based PP operations.

Gathering phantom layers from all ranks in the forward propagation corresponds to an \texttt{All-Gather} operation, which collects and concatenates data from all ranks in every rank. Likewise, during backpropagation, the partial gradients on each GPU need to be aggregated and distributed to all ranks, which corresponds to a \texttt{Reduce-Scatter} operation. However, PyTorch does not provide built-in layers that support these PP-specific collective operations for both forward and backward propagation in a distributed setting. To address this issue, we implemented custom gradient computations by extending the \texttt{torch.autograd.Function} class. This class allows overriding the `forward` and `backward` methods with custom logic, as outlined in Algorithm~\ref{alg:custom_autograd}, enabling communication between compressors and decompressors. An extension of the \texttt{torch.autograd.Function} class, called \textsc{AllGatherFunction}, was implemented using PyTorch’s \texttt{dist.all\_gather} and 
\texttt{dist.reduce\_scatter} collectives. 
\section{Performance Results}
\label{sec:perf}
We present results that compare the performances of TP and PP for {\bf fixed number of epochs } in Sec.~\ref{sec:parperf} and for {\bf fixed loss} in Sec.~\ref{sec:energyperf}. The first allows parallel performance comparisons based on their execution characteristics, and the second compares their respective energy consumption when trained to a fixed model loss. 

%\subsection{Data and Hardware}
\noindent{\em Data and Hardware:}
The data set used in the experiments to train the TP and PP models was created using a standard Gaussian matrix, $W \in \mathbb{R}^{n \times n}$, that was kept fixed for all the examples. The data set was generated as $\{(x_i,y_i): i=1,\dots N\}$ pairs where $x_i, y_i\in \mathbb{R}^n$ and $y_i = \sigma(W \sigma(x_i))$ with $\sigma =$ ReLu.

%The reported training experiments were performed on a data set $\{(x_i,y(x_i)): i=1,\dots N\}$ with $N=10000$. The base function was $y(x) = \sigma(W \sigma(x))$ where $\sigma =$ReLu and $W \in \mathbb{R}^{n \times n}$ is a realization of standard Gaussian matrix that was kept fixed throughout all the examples. 

The \textsc{Frontier} supercomputer was used for this study. It is a HPE Cray supercomputer with AMD compute nodes, each featuring a 64-core “Optimized 3rd Gen EPYC” CPU and 512 GB of DDR4 memory. Every node hosts four AMD MI250X accelerators, each with two Graphics Compute Dies (GCDs), together offering eight GCDs per node with 64 GB of high-bandwidth memory (HBM2E) each. Nodes are connected by AMD’s Infinity Fabric, enabling up to 36 GB/s bidirectional bandwidth between CPU and GPU, 200 GB/s peak bandwidth between GCDs on the same MI250X, and 50–100 GB/s between GCDs located on different MI250X accelerators.

%\textcolor{blue}{To measure energy consumption by different parallel training strategies, we developed a script that can run in the background along with the main training job and read power and energy statistics from the sensors attached to the GPU cards. Our script runs in the background and utilizes AMD's rocm-smi tool to measure power at a regular interval. While for distributed training in Tensor and Phantom parallel strategies we launch one process per GCD, we launch one energy measurement script per node. This is because rocm-smi can read sensor data from all GPUs in a node. There are a total of eight GCDs in a single Frontier node, packed in four MI250X GPU cards, two GCDs in a single GPU card. During the measurement of energy, we observed a significant lead time in preparation of the data, model, and hardware warm-up. For the small problem sizes used in our energy comparison experiments, actual training time is a fraction of the total run-time. So, to calculate the total energy consumption by the training, we take the area under the power curve (power vs time) over the actual training duration.}

\subsection{Parallel Execution Performance} %(Fixed Number of Epochs)}
\label{sec:parperf}
Relative performances are compared with respect to fixed values of $n$, $L$ and $p$, unless otherwise specified. In addition, identical training data sets and hyperparameters, such as optimizers, batch sizes, etc., were used for the comparisons. In addition, the parallel execution performance of TP and PP approaches when training a FNN are compared for a {\em fixed number of iterations (epochs)}. 

\begin{figure*}[thbp]
    \centering
    \subfloat[\scriptsize $n = 65,536$, $L=6$ and $k=64$.]{
        \includegraphics[width=0.32\textwidth]{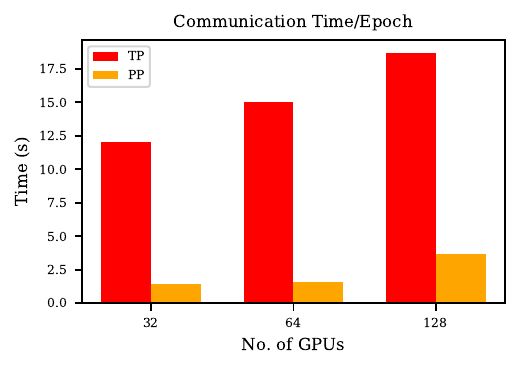}
        \label{fig:comm_time_n65536}
    }
    %\hfill
    \subfloat[\scriptsize $n = 4,096$ and $L=2$.]{
        \includegraphics[width=0.32\textwidth]{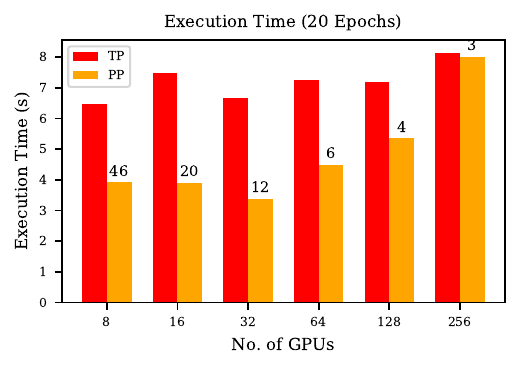}
        \label{fig:exec_time_n4096}
    }
    %\hfill
    \subfloat[\scriptsize $n = 16,384$ and $L=2$.]{
        \includegraphics[width=0.32\textwidth]{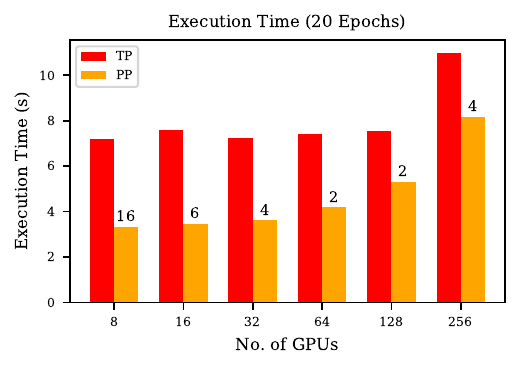}
        \label{fig:exec_time_n16384}
    }
    \caption{Comparison of TP and PP performance across model sizes and GPU counts. The phantom layer sizes (numbers of ghost neurons), $k$, is indicated in (b) and (c) for PP.}
    \label{fig:tp_pp_subfigs}
\end{figure*}

\fig{comm_time_n65536} illustrates the relative communication overheads of TP and PP when training a FNN of width $n=65,536$ and $L=6$ to a fixed number of epochs. For PP training, the number of ghost neurons (width of the phantom layer) was chosen as $k=64$. The net communication cost per epoch (iteration) for PP training is seen to be significantly lower than that of TP training across 32, 64 and 128 GPUs due to the larger bandwidth-bound communication costs of TP compared to PP, as indicated by \eqn{tpcomm} and \eqn{ppcomm}.

\fig{exec_time_n4096} shows the total execution time (communication and computation) for a small two-layer FFN model with $n=4096$. The PP executions are labeled by the size of the phantom layer. The total execution time per epoch for PP are shown to outperform TP across a wide range of GPU counts. However, as the number of GPUs increases, the relative performance tends to converge. This convergence is due to the small FFN size which makes the overall execution communication bound, and the bandwidth-bound communication costs of both approaches become comparable. As the size of the model increases, PP regains its advantage over TP as shown in \fig{exec_time_n16384} for a two-layer FFN model with $n=16,384$ with PP outperforming TP at fixed $p$ even with larger phantom layers (e.g., $k=4$ in \fig{exec_time_n16384} while $k=3$ in \fig{exec_time_n4096} for $p=256$).

\begin{figure}[thbp]
    \centering
    \includegraphics[scale=0.8]{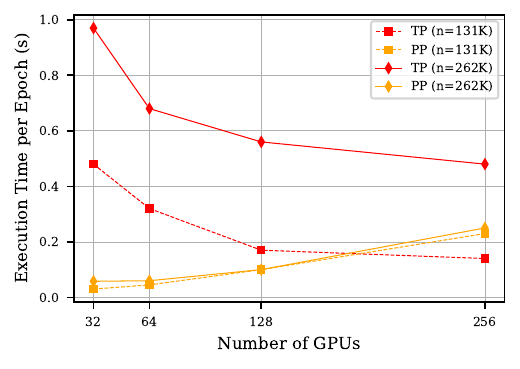}
    \caption{Comparison of TP and PP for large model sizes.}
    \label{fig:large-ffn}
\end{figure}

Next, we evaluate the performance of TP and PP for large model sizes. We consider input sizes of $n = 2^{17} = 131,072$ and $n = 2^{18} = 262,144$. For both cases, $k$ is kept fixed at 64. The number of GPUs is varied from 32 to 256. \fig{large-ffn} reports the execution time per epoch vs. the number of GPUs. To ensure accurate timing, the first epoch is excluded from the measurements for both TP and PP, as PyTorch incurs significant overhead during initial data structure creation.

Across all configurations, PP shows lower memory consumption compared to TP, which is by design (see \eqn{tpcomp} and \eqn{ppcomp}), but the relative memory advantage is magnified for large $n>kp$. For $n = 131,072$, PP consistently outperforms TP up to $p=128$, as shown in \fig{large-ffn}. However, at $p=256$, TP overtakes PP. This performance flip-flop is attributed to an increase in PP overhead from the management of additional data structures required for gradient aggregation which is proportional to $p$, as discussed below.

The origins of this flip-flop can be traced back to the performance of GEMM operations. Computations of $\mathbf{L}_l^{(i)}$ on rank $i$ are large GEMM operations with dimensions $(n/p) \times (n/p)$ by $(n/p) \times \textit{batch\_size}$ while the $(p - 1)$ decompression operations, $\mathbf{D}_l^{(j,i)}$), are smaller GEMM operations with dimensions $k \times \textit{batch\_size}$ by $\textit{batch\_size} \times (n/p)$. As noted in~\cite{nvidia-gemm-perf}, the performance of GEMM decreases with smaller problem sizes, making the decompressor less efficient due to its relatively small $k$ dimension. Moreover, because the number of decompressors grows with $p$, its total execution time increases nearly linearly with the GPU count.
%the number of GPUs.  

For the larger FFN in \fig{large-ffn} with $n = 262,144$, PP maintains superior performance across all tested GPU counts. Notably, TP could not be executed on $p=32$ due to memory exhaustion and is therefore omitted from the corresponding results. In contrast, the reduced memory footprint of PP enables successful training even at this lower scale. Furthermore, PP continues to outperform PP at $p=256$, highlighting its scalability advantages. These results suggest that with sufficient computing resources, PP not only achieves better run-time performance but also offers improved memory efficiency. %Potential energy savings from PP compared to TP are discussed next.

\subsection{Energy Consumption} %(Fixed Model Loss)}
\label{sec:energyperf}
The results in the previous section empirically validate \eqn{pplttp} but {\bf does not} imply that $E_\pi(n,p,k,L) < E_\tau(n,p,L)$ since the number of PP and TP iterations ($\nu_\pi$ and $\nu_\tau$, respectively) needed to train a model to a {\em fixed} loss, $\lambda$, are not guaranteed to be equal (see \eqn{totale}). Here, we explore the total energy consumed by the two parallel approaches to reach a {\em fixed} loss.

Recall that a PP model with $k$ ghost neurons is guaranteed to be smaller than the corresponding TP model when $k<n/p$ (see \eqn{kbound}) which implies that, compared to a TP mode, a PP model:
\begin{itemize}
    \item has a smaller memory requirement as already reported in the previous section.
    \item is likely to require fewer iterations to train to the same loss since there are fewer parameters to train. 
\end{itemize}

To test the second possibility, the baseline FFNs were first trained to a fixed loss using TP, then using PP with different phantom layer widths, $k$, for a given $p$ and repeated over different values of $p$. For each TP and PP execution, the number of iterations, $\nu_\tau$ and $\nu_\pi$, to reach the same fixed loss was recorded. In the absence of communication overheads, the total computation cost scales with the model size, which implies that the product of the iteration count needed to reach the target loss and the model size is expected to scale with the net energy consumed to train the model to that loss.

\begin{figure*}[thbp]
    \centering
    \subfloat[]{
        \includegraphics[width=0.32\textwidth]{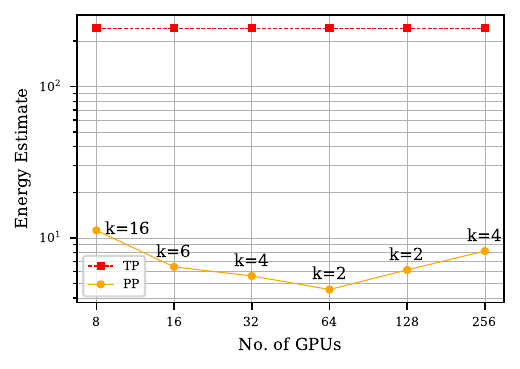}
        \label{fig:energy_est_n16384}
    }
    %\hfill
    \subfloat[]{
        \includegraphics[width=0.32\textwidth]{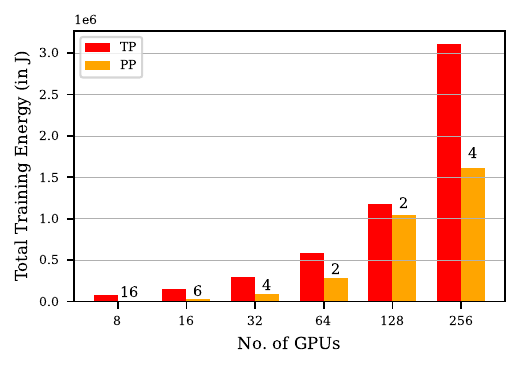}
        \label{fig:energy_n16384}
    }
        %\hfill
    \subfloat[]{
        \includegraphics[width=0.32\textwidth]{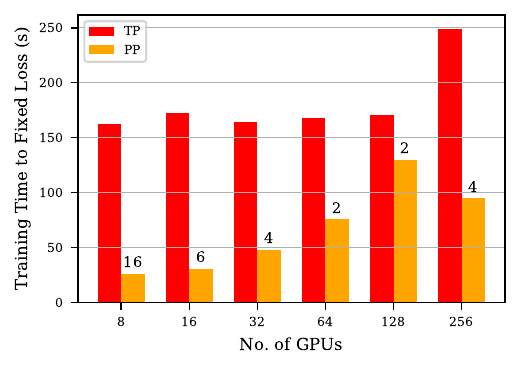}
        \label{fig:time_n16384}
    }
    \caption{Energy consumption and execution time to train a FFN with $n = 16,384$ and $L=2$ to the {\bf same (fixed) loss} using TP and PP.}
    \label{fig:energy_subfigs}
\end{figure*}

These energy estimates are presented in \fig{energy_est_n16384} for a FFN with $n=16,384$ and $L=2$. Each model in \fig{energy_est_n16384} was trained to exactly the same (fixed) loss. Recall that the size of a TP model remains unchanged with $p$; only the overall computations, which scale with the model size, are partitioned across the $p$ ranks. Consequently, the iteration count to train a TP model to a fixed loss also remains independent of $p$ (see Table~\ref{tab:energy_comparison_n16384_with_model_size}). Therefore, a communication-free estimate of the TP energy consumption remains unchanged when varying $p$ as shown in \fig{energy_est_n16384} (red).

On the other hand, the size of a PP model depends on $p$ and $k$ (see \eqn{ppcomp}). PP models trained with various $k$ and $p$ will result in different model sizes and different iteration counts to reach a target loss, as shown in Table~\ref{tab:energy_comparison_n16384_with_model_size}. Each PP model can be seen to be smaller than the TP model across a range of $p$ and the PP iteration counts needed to train these smaller PP models are accordingly seen to be smaller than the TP iteration counts. Since an PP model is smaller than a corresponding TP model by design, a communication-free energy estimate of a PP model that takes fewer iterations to reach the target loss is expected to be lower than that of a TP model for the same $p$. \fig{energy_est_n16384} shows these PP energy estimates across a range of values of $k$ and $p$.  

\begin{table*}[th]
\centering
\small
\begin{tabular}{c c
                c c c c
                c c c c}
\toprule
    &  & \multicolumn{4}{c}{\textbf{Tensor Parallel}} & \multicolumn{4}{c}{\textbf{Phantom Parallel}} \\
\cmidrule(lr){3-6} \cmidrule(lr){7-10}
$p$    & $k$ & \makecell{Model Size \\ (in M)} & \makecell{Energy/Epoch\\(in J)} & \makecell{No. of\\Epochs} & \makecell{Total Energy\\(in J)} 
          & \makecell{Model Size \\ (in M)} & \makecell{Energy/Epoch\\(in J)} & \makecell{No. of\\Epochs} & \makecell{Total Energy\\(in J)} \\
\midrule
8   & 16 & 537 & 181.2   & 453 & 82,084     & 71 & 74.6    & 157 & 11,712     \\
16  & 6  & 537 & 333.6   & 453 & 151,121    & 37 & 158.0   & 175 & 27,650     \\
32  & 4  & 537 & 658.4   & 453 & 298,255    & 21 & 333.6   & 267 & 89,071     \\
64  & 2  & 537 & 1281.6  & 453 & 580,565    & 13 & 779.2   & 362 & 282,070    \\
128 & 2  & 537 & 2611.2  & 453 & 1,182,874  & 13 & 2131.2  & 488 & 1,040,026  \\
256 & 4  & 537 & 6873.6  & 453 & 3,113,741  & 36 & 6950.4  & 232 & 1,612,493  \\
\bottomrule
\end{tabular}
\caption{Comparison of energy consumption using TP and PP to train a FFN with $n = 16,384$ and $L=2$ to a fixed MSE loss.}
\label{tab:energy_comparison_n16384_with_model_size}
\end{table*}

Since it was analytically proven that the PP communication cost per iteration is always less than the TP communication cost per iteration for fixed $n$, $p$, $L$ and target loss $\lambda$ when $k<n/p$ (see \eqn{commineq}) and demonstrated empirically in the previous section (see \fig{comm_time_n65536}), the energy consumption of the PP models shown in \fig{energy_est_n16384} are also expected to be smaller than that of the corresponding TP models even in the presence of inter-process communications. This hypothesis was tested by measuring the energy consumption (in Joules) when training the TP and PP models in \fig{energy_est_n16384}.

To measure and quantify energy consumption, a background monitoring script that operates concurrently with a training process was developed to sample power and energy statistics from GPU-attached sensors. The script leverages AMD’s \textsc{rocm-smi} utility to collect power measurements at fixed intervals. In distributed training configurations, one training process per GCD (Graphics Compute Die) was instantiated. For energy monitoring, a single instance of the measurement script was launched per node, as \textsc{rocm-smi} provides visibility into all GPUs within a node. 
%Each \textsc{Frontier} node comprises eight GCDs, integrated across four MI250X GPU cards, with two GCDs per card.

Empirical observations revealed an initialization phase involving data loading, model construction and hardware warm-up, which introduced non-negligible lead time prior to the commencement of training. Given the relatively small problem sizes used in this study for energy comparisons, the training phase constituted a relatively smaller portion of the total execution time. In real-world training campaigns with super-sized transformer-based or frontier models, the total execution time is overwhelmingly dominated by the training time. As such, to isolate training-specific energy consumption, the area under the power-time curve over the precise duration of the training phase was computed, excluding initialization overheads from the energy accounting. 

This script was used to perform measurements of the energy consumption when training each model in \fig{energy_est_n16384}. The results are plotted in \fig{energy_n16384} and listed in Table~\ref{tab:energy_comparison_n16384_with_model_size}. These empirical results demonstrate that training a PP model is more energy-efficient than training the corresponding TP model on the same number of ranks. For $p=256$, the energy consumption of PP training is $\sim 50\%$ of TP training. 

These results open up the possibility of even greater savings when the two approaches are compared for different values of $p$. \fig{time_n16384} shows the wall-time to train the FFN using TP and PP with different phantom layer sizes to a fixed loss on 8 to 256 GPUs. The increase in wall-time for PP as $p$ grows was already discussed in Sec.~\ref{sec:parperf} (see \fig{large-ffn}). Despite this trend, the PP execution times outperform TP execution times. Notably, the PP execution time on $p=8$ is significantly smaller than for TP (and PP with $k=4$) on $p=256$. This is because of the smaller PP model size and fewer iterations on $p=8$ with $k=16$ in addition to smaller communication costs at $p=8$ than at $p=256$. Using Table~\ref{tab:energy_comparison_n16384_with_model_size}, a comparison of the energy consumption of training the $537M$ TP model on $p=256$ with training the $71M$ PP model on $p=8$ to the {\em same} target loss yields over two orders of magnitude lesser energy consumption in PP training compared to TP training while registering an order of magnitude reduction in the training time.
\section{Conclusions and Future Work}
\label{sec:conclusions}
The paper presented a comprehensive proof-of-principle study of \textit{phantom parallelism}, a proposed alternative to tensor parallelism aimed at reducing energy consumption in large-scale neural network training workloads. The study presented here was confined to the simpler FFN architecture due to the ease of implementing and testing the central principles of the proposed approach. By introducing a phantom projection of intermediate activations, the dominant communication bottleneck in tensor parallelism was shown to be significantly mitigated without compromising model performance. The paper presented thorough derivations of non-trivial forward and backward propagation operators integrated as custom PyTorch autograd operators. Experiments on up to 256 GPUs were shown to deliver up to 50\% reduction in the energy consumed to train FFNs using the proposed phantom parallel approach when compared with conventional tensor parallel methods. The proposed approach was also shown to train smaller phantom models on smaller GPU counts to the same model loss as larger tensor parallel models on larger GPU counts offering the possibility for even greater energy savings.

%Though limited to FFNs, the study suggests that phantom parallelism offers a viable and energy-efficient alternative to tensor parallelism. Phantom parallelism preserves compatibility with existing optimizers, data loaders, and training pipelines. These characteristics make it a promising replacement for tensor parallelism in bandwidth- or power-constrained training environments. Future work will focus on generalizing phantom parallelism to full transformer architectures, extending it to inference workloads, and integrating it with pipeline and data parallelism for broader deployment in next-generation training systems for super-sized models.

Though limited to FFNs, future work will focus on generalizing phantom parallelism to more complex neural network architectures such as the full transformer architecture. Each transformer block consists of a self-attention sub-block followed by a FFN sub-block. In the self-attention sub-block, the dominant operation involves multiplying a square weight matrix $\mathbf{W} \in \mathbb{R}^{d \times d}$ with a tall-skinny matrix $\mathbf{H} \in \mathbb{R}^{d \times t}$, where $t$ is the number of tokens, $d$ is the embedding dimension and $t \ll d$. Since the FFN computations are also dominated by square matrix--vector operations, the phantom parallel substitution extends naturally to self-attention: $\mathbf{H}$ can be interpreted as a collection of $t$ column vectors, each processed independently using the same phantom parallel strategy. Consequently, the communication-to-computation ratio for self-attention is asymptotically identical to that for the FFN, as both computation and communication costs scale linearly with $t$. Note that $d$ in this section corresponds to $n$ in the rest of the paper.

The comparisons in this paper did not include the effects of optimization strategies such as pruning, quantization, or mixed precision. These approaches are complementary and expected to apply equally to TP and PP. However, because PP inherently reduces the number of trainable weights when $k < n/p$ (see \eqn{pplttp}), we expect these optimizations to be more favorable to PP. 
%Even without optimizations, the results presented in this paper demonstrate that PP outperforms TP in both communication and training time. 
Future work will compare and study the effects of such external optimization strategies on performances of TP and PP training pipelines in the context of simpler FFNs as well as more complex transformer architectures.
\setlength{\abovedisplayskip}{4pt}
\setlength{\belowdisplayskip}{4pt}
\setlength{\abovedisplayshortskip}{2pt}
\setlength{\belowdisplayshortskip}{2pt}
\setlength{\textfloatsep}{6pt}
\setlength{\floatsep}{6pt}
\renewcommand{\arraystretch}{1.0}

\appendix 
\section{Appendix}
\label{sec:comm}
Table~\ref{tab:commstppp} lists the collective communication calls per layer per iteration of a TP and PP execution using $p$ processes. For details of collective communications used in TP, see~\cite{nvidia-megatron-tp,pytorch-colwise-parallel}. Details of the collective communications used in PP executions are presented in Sec.~\ref{sec:ppoperators}.
\begin{table}[ht]
\centering
\caption{Communications in tensor and phantom parallelism}
\small
\begin{tabular}{llcl}
\toprule
\textbf{Model} & \textbf{Collective} & \textbf{Message Size, $m$} & \textbf{Direction} \\
\midrule
TP & \texttt{Broadcast}       & $n\times$ batch\_size      & Forward    \\
TP & \texttt{All-Gather}      & $\tfrac{n}{p}\times$ batch\_size    & Forward    \\
TP & \texttt{All-Reduce}      & $n\times$ batch\_size      & Backward   \\
TP & \texttt{Reduce-Scatter}  & $\tfrac{n}{p}\times$ batch\_size    & Backward   \\
\midrule
PP & \texttt{All-Gather}      & $k\times$ batch\_size      & Forward    \\
PP & \texttt{Reduce-Scatter}  & $k\times$ batch\_size      & Backward   \\
\bottomrule
\end{tabular}
\label{tab:commstppp}
\end{table}
Note that the extra \texttt{Broadcast} and \texttt{All-Reduce} collective communications are necessary in a TP execution because the global layer is required on each rank to ensure correctness of the forward computations and backward gradient computations.   

We evaluated the performance scaling of \texttt{All-Reduce}, \texttt{Broadcast}, \texttt{Reduce-Scatter} and \texttt{All-Gather} collectives on the \textsc{Frontier} supercomputer by fitting empirical measurements to a unified communication model of the form:
\bea
\label{eqn:ppcollective}
%\log_2(\text{Time}) = \log_2(\alpha \log_2 P + \beta \cdot \tfrac{n}{P} + c),
%comm\_time(n,p) = c_1\log_2 p + c_2 \cdot \tfrac{n}{p} + c_3
comm\_time(m,p) = c_1\log_2 p + c_2 \cdot m + c_3
\eea
which captures both latency-bound and bandwidth-bound communication costs involving $p$ processes and message size $m$. Table~\ref{tab:comm-models} lists the fitted parameters for each collective. Empirical data was collected for message sizes $m$ ranging from $2^2$ to $2^{26}$ floats and $p$ ranging from 2 to 256 GPUs. 
\begin{table}[ht]
\small
\setlength{\tabcolsep}{4pt}
\centering
\caption{Communication model for collective operations.}
\begin{tabular}{lccc}
\toprule
\textbf{Collective} & $c_1$ & $c_2$ & RMSE \\
\textbf {Communication} & (Latency) & (Bandwidth) & [$\log_2(\mu s)$] \\
\midrule
%{\bf Regular} &  &  &  \\
\texttt{Broadcast} & $35.5$ & $1.12 \!\times\! 10^{-3}$ & 3.20 \\
\texttt{All-Reduce} & $33.4$ & $2.56 \!\times\! 10^{-3}$ & 2.58 \\
%{\bf Phantom} &  &  &  \\
\texttt{All-Gather} & $149.94$ & $2.07 \!\times\! 10^{-3}$ & 3.90 \\
\texttt{Reduce-Scatter} & $145.52$ & $2.40 \!\times\! 10^{-3}$ & 3.91 \\
\bottomrule
\end{tabular}
\label{tab:comm-models}
\end{table}
As an example, for \texttt{Reduce-Scatter}, the fitted parameters were $c_1 = 145.5$, $c_2 = 2.40 \times 10^{-3}$, $c_3 = 2.62 \times 10^{-58}$ and RMSE = 3.91 in $\log_2(\mu s)$ (in other words, an average deviation of $2^{3.9}\sim 15\mu s$). We ignore the overhead $c_3$ since $c_3\sim 0$ for all the collective communications in both types of execution. 
%The behavior of \texttt{all-gather} yielded $c_1 = 149.9$, $c_2 = 2.07 \times 10^{-3}$, and RMSE = 3.90. These results indicate that both collectives operate within a consistent latency-bandwidth regime, with effective per-node bandwidth exceeding 120~GB/s. 
Due to the close fit of the empirical data to the analytic model as indicated by the low residuals, 
%suggest that tree-based collective implementations are highly optimized on Frontier’s Slingshot interconnect, 
we use \eqn{ppcollective}, with the appropriate constants in Table~\ref{tab:comm-models}, to model the per-iteration communication cost in both types of parallel execution.

%\iffalse
\section*{Acknowledgment}
  This manuscript has been authored by UT-Battelle, LLC under Contract No.
  DE-AC05-00OR22725 with the U.S. Department of Energy. The United States
  Government retains and the publisher, by accepting the article for publication,
  acknowledges that the United States Government retains a non-exclusive, paid-up,
  irrevocable, worldwide license to publish or reproduce the published form of
  this manuscript, or allow others to do so, for United States Government
  purposes. The Department of Energy will provide public access to these results
  of federally sponsored research in accordance with the DOE Public Access Plan
  (http://energy.gov/downloads/doe-public-access-plan). This research was sponsored by Oak Ridge National Laboratory’s Laboratory Directed Research and Development program. This research used resources at the Oak Ridge Leadership Computing Facility which is a DOE  Office of Science User Facility.
%\fi 

\iffalse
\begin{table}[h]
\centering
\caption{FY26 Effort and Budget Allocation}
\begin{tabular}{@{}lll@{}}
\toprule
\textbf{Name} & \textbf{FTE \%} & \textbf{FY26 (\$)} \\
\midrule
Seal    & 0.25 & 162{,}843 \\
Ramirez & 0.15 & 85{,}067 \\
Alam    & 0.15 & 71{,}248 \\
Lu      & 0.10 & 49{,}188 \\
Dash    & 0.10 & 39{,}976 \\
Lunga   & 0.05 & 32{,}569 \\
Travel   & --  & 9,100 \\
\midrule 
\textbf{Total}   &  & 449,991 \\
\bottomrule
\end{tabular}
\end{table}
\fi 

\printbibliography
\end{document}